\documentclass[10pt, a4paper]{article}
\usepackage{lrec}
\usepackage{multibib}
\newcites{languageresource}{Language Resources}
\usepackage{graphicx}
\usepackage{tabularx}
\usepackage{soul}
\usepackage{booktabs} 
\usepackage{nicefrac}
\usepackage{latexsym}
\usepackage{multirow}
\usepackage{url}
\usepackage{csquotes}
\usepackage{array}
\usepackage{ctable}
\usepackage{epstopdf}
\usepackage[latin1]{inputenc}
\usepackage{hyperref}
\usepackage{xstring}
\newcolumntype{P}[1]{>{\centeringarraybackslash}p{#1}}

\title{Network Features Based Co-hyponymy Detection}

\name{Abhik Jana, Pawan Goyal}

\address{Indian Institute of Technology Kharagpur, Indian Institute of Technology Kharagpur \\
        abhik.jana@iitkgp.ac.in, pawang@cse.iitkgp.ac.in\\}

\abstract{
Distinguishing lexical relations has been a long term pursuit in natural language processing (NLP) domain. Recently, in order to detect lexical relations like hypernymy, meronymy, co-hyponymy etc., distributional semantic models are being used extensively in some form or the other. Even though a lot of efforts have been made for detecting hypernymy relation, the problem of co-hyponymy detection has been rarely investigated. In this paper, we are proposing a novel supervised model where various network measures have been utilized to identify co-hyponymy relation with high accuracy performing better or at par with the state-of-the-art models.       
\\ \newline  }

\begin{document}

\maketitleabstract

\section{Introduction}
Automatic detection of lexical relations is a fundamental task for natural language processing (NLP). Numerous applications including paraphrasing, query expansion, recognizing textual entailment, ontology building, metaphor detection etc. are benefited by precise relation classification and relation discovery tasks. 
For example, it may be difficult to interpret a sentence containing a metaphor, like ``He drowned in a sea of grief'' if we go by the literal meaning. But if we replace `drowned' by its co-hyponym `overwhelmed' and `sea' by its co-hyponym `lot', it immediately provides an inference. Note that, `drown' and `overwhelm' are (co-)hyponyms for the concept `cover' whereas `sea' and `lot' are (co-)hyponyms for the concept `large indefinite amount' as per WordNet~\cite{miller1995wordnet}.
\if{0}
For example, it may be difficult to interpret a query ``cat noseprint". But if we replace this pair by ``human fingerprint", it immediately provides an inference. Note that, we have just replaced the terms with their co-hyponyms; `cat' and `human' are co-hyponyms for the concepts `animal' whereas `noseprint' and `fingerprint' are co-hyponyms for the concept `biometric authentication'.  \textbf{TODO: Can be have a better motivation?}  
\fi

Lexical relations are of variety of types like hyponyms, hypernyms, co-hyponyms, meronyms etc. Among these, some relations are symmetric (co-hyponymy) and some are asymmetric (hypernymy, meronymy). With the advancement of distributional semantics representation of words, researchers have attempted to identify lexical relations in both supervised and unsupervised ways.

One of the oldest attempt for detection of hypernymy extraction dealt with finding out `lexico-syntactic patterns' proposed by~\newcite{hearst1992automatic}. A lot of attempts have been made for hypernymy extraction using knowledge bases like Wordnet, Wikipedia and hand crafted patterns or patterns learnt from the corpus~\cite{cederberg2003using,yamada2009hypernym}. With the emergence of the trend of applying distributional hypothesis~\cite{firth1957synopsis} to solve this relation classification task, researchers have started using Distributional Semantic Models (DSM) and have come up with several directional measures~\cite{roller2014inclusive,weeds2014learning,SANTUS16.455,shwartz-santus-schlechtweg:2017:EACLlong,roller-erk:2016:EMNLP2016}. Specifically for hypernymy detection, researchers also used a variant of distributional hypothesis, i.e., distributional inclusion hypothesis~\cite{geffet2005distributional} according to which the contexts of a narrow term are also shared by the broad term. Recently, entropy-based distributional measure~\cite{santus2014chasing} has also been tried out for the same purpose. In some of the recent attempts~\cite{fu2014learning,yu2015learning,nguyen-EtAl:2017:EMNLP2017}, people have tried several embedding schemes for hypernymy detection. One interesting attempt was made by~\newcite{kiela2015exploiting}, where they exploited image generality for lexical entailment detection.
Most of the attempts made for meronymy detection are mainly pattern based~\cite{berland1999finding,girju2006automatic,pantel2006espresso}. Later, investigations have been made for the possibility of using distributional semantic models for part-of relations detection~\cite{morlane2015can}. As far as co-hyponymy detection is concerned, researchers have tried with several DSMs and measures for distinguishing hypernyms from co-hyponyms but the number of attempts is very small. One such attempt is made by~\newcite{weeds2014learning}, where they proposed a supervised framework and used several vector operations as features for the classification of hypernymy and co-hyponymy. In one of the recent work~\cite{SANTUS16.455}, a supervised method based on a Random Forest algorithm has been proposed to learn taxonomical semantic relations and they have shown that the model performs good for co-hyponymy detection.   

It is evident from the literature that, most of the efforts are made for hypernymy or lexical entailment detection; very few attempts have been made for co-hyponymy detection. In this paper, we are proposing a supervised framework for co-hyponymy detection where complex network measures are used as features. Network science has always been proved to be very effective in addressing problems including the structure and dynamics of the human brain, the functions of genetic pathways, social behavior of humans in the online and offline world. Researchers have tried to understand human language using complex network concepts as well~\cite{antiqueira2007strong,ferrer2007spectral}. Many works like co-occurrence network~\cite{i2001small}, syntactic dependency network~\cite{i2004r} etc. exist where network properties are applied to natural language processing tasks, which lead to elegant solutions to the problem. These works constitute our prime motivation to apply network science methods for co-hyponymy detection. 

\noindent {\bf Network features:} In particular, we propose a supervised method based on the theories of complex networks to accurately detect co-hyponymy relationship. Our study is based on a unique network representation of the corpus called a distributional thesauri (DT) network~\cite{riedl2013scaling} built using Google books syntactic n-grams. We hypothesize that, if two words are having `co-hyponymy' relationship, then those words are distributionally more similar compared to the words having hypernymy, meronymy relationship or any random pair of words. In order to capture the distributional similarity between two words in the DT network, we are proposing the following five network measures for each word pair: (i) structural similarity ($SS$), (ii) shortest path ($SP$), (iii) weighted shortest path ($SPW$), (iv) edge density among the intersection of neighborhoods($ED_{in}$), (v) edge density among the union of neighborhoods ($ED_{un}$). A remarkable observation is that although this is a small set of only five features, they are able to successfully discriminate co-hyponymy from hypernymy, meronymy and random pairs with high accuracy.

\noindent {\bf Classification model:} We use these five network measures as features to train classifiers like SVM, Random Forest to distinguish the word pairs having co-hyponymy relation from the word pairs having hypernymy or meronymy relation, or from any random pair of words. 

\noindent {\bf Evaluation results:} We evaluate our approach by three experiments. In the first two experiments, taking two different baselines~\cite{weeds2014learning,SANTUS16.455}, we follow their experimental setup as well as their publicly available dataset and show that using our proposed network features, we are able to improve the accuracy of the co-hyponymy detection task. In the third experiment, we prepare three datasets extracted from BLESS dataset~\cite{baroni2011we} for three binary classification tasks: Co-hyponymy vs Random, Co-hyponymy vs Meronymy, Co-hyponymy vs Hypernymy and show that we get consistent performance as the previous two experiments, achieving accuracy in the range of 0.73-0.97. We have made these three datasets publicly available\footnote{\url{http://tinyurl.com/y99wfhzb}}. 

\section{Methodology}
As a graph representation of words, we use distributional thesauri (DT) network~\cite{riedl2013scaling} from the Google books syntactic n-grams data~\cite{goldberg2013dataset} spanning from 1520 to 2008. In a graph structure, the DT contains for each word a list of words that are similar with respect to their bi-gram distribution~\cite{riedl2013scaling}. 
\begin{figure}[!h]
\begin{center}
\includegraphics[scale=0.65]{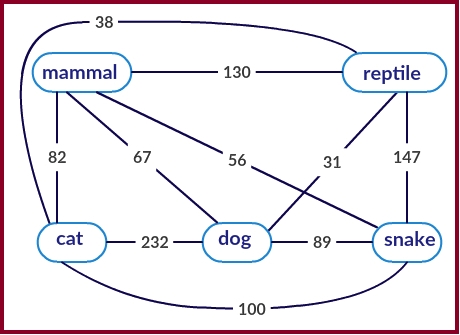} 
\caption{A sample snapshot of Distributional Thesaurus Network where each node represents a word and the weight of edge between two words is defined as the number of context features that these two words share in common. Here the word `cat' shares more context features with its co-hyponym `dog' compared to their common hypernym `mammal'.}
\label{DT}
\end{center}
\vspace{-0.5cm}
\end{figure}

\begin{table*}[!tbh]
\small

\begin{center}
    \begin{tabular}{|>{\centering}p{2cm}|>{\centering}p{3cm}|>{\centering}p{1.1cm}|>{\centering}p{1.1cm}|>{\centering}p{1.1cm}|>{\centering}p{1.1cm}|>{\centering}p{1.1cm}|}
    \hline
\bf{Type} &	\bf{Word pair} & \boldmath{$SS$} & \boldmath{$SP$} & \boldmath{$SPW$} & \boldmath{$ED_{in}$} & \boldmath{$ED_{un}$} \tabularnewline \hline
co-hyponymy & snake - crocodile	& 0.7 &  1	& 0.84	&	0.57	& 0.36 \tabularnewline \hline 
hypernymy 	& snake - reptile &	0.67 & 1 & 0.85 & 0.5	& 0.31 \tabularnewline \hline
meronymy 	& snake - scale &	0 & 2 & 1.99 & 0	& 0.15 \tabularnewline \hline
random 	& snake - permission &	0 & 3 & 2.98 & 0	& 0.14 \tabularnewline \hline
    \end{tabular}
\end{center}
\vspace{-0.4cm}
	\caption{The network properties of sample cases taken from BLESS dataset.}
\vspace{-0.5cm}
\label{samples}
\end{table*}

In the network, each word is a node and there is a weighted edge between a pair of words where the weight of the edge is defined as the number of features that these two words share in common. A snapshot of the DT is shown in Figure~\ref{DT}. Our hypothesis is that the word pairs having co-hyponymy relation are distributionally more similar than the words having hypernymy or meronymy relation or any random pair of words. Now, if two words are distributionally similar, it will be reflected in the DT network in that they will exist in close proximity, their neighborhood will contain similar nodes and the connections among their neighborhood will be dense. In order to capture the notion of distributional similarity among the word pairs, we choose five \textit{cohesion indicating} network properties: (i) structural similarity ($SS$), (ii) shortest path ($SP$), (iii) weighted shortest path ($SPW$), (iv) edge density among the intersection of neighborhoods ($ED_{in}$), (v) edge density among the union of neighborhoods ($ED_{un}$).

Let $(w_i, w_j)$ be the pair of words for which we compute the following network measures -

{\bf Structural Similarity (SS):} The structural similarity $SS(w_i,w_j)$ is computed as:
\begin{equation}
SS(w_i,w_j) = \frac{N_c}{\sqrt{deg(w_i)*deg(w_j)}}
\label{Structural_Similarity}
\end{equation}
where $N_c$ denotes the number of common neighbors of $w_i$ and $w_j$ and $deg(w_k)$ denotes the degree of $w_k$ in the DT graph, for $k=i,j$. \\

{\bf Shortest Path (SP):}  This is a measure of distance of the shortest path between $w_i$ and $w_j$ in DT network. \\

{\bf Weighted Shortest Path (SPW):}  
The weighted shortest path $SPW(w_i,w_j)$ is computed as: 
\begin{equation}
SPW(w_i,w_j) = SP(w_i,w_j)-\frac{EW_{average}}{EW_{max}}
\label{SPW}
\end{equation}
where $SP(w_i,w_j)$ gives the length of the shortest path between $w_i$ and $w_j$; $EW_{average}$ gives the average edge weight along the shortest path; $EW_{max}$ gives the maximum edge weight in the DT network, which is 1000 in our case. \\

{\bf Edge density among the intersection of neighborhoods ($ED_{in}$):} 
\begin{equation}
ED_{in}(w_i,w_j)= \#(A_{in})/ \#(P_{in})
\label{EDIN}
\end{equation}
where $A_{in}$ denotes the actual edges present between the common neighbors of $w_i$ and $w_j$ and $P_{in}$ denotes the maximum possible edges between the common neighbors, i.e., $\frac{n(n-1)}{2}$. \\

{\bf Edge density among the union of neighborhoods ($ED_{un}$):} 
\begin{equation}
ED_{un}(w_i,w_j)= \#(A_{un})/ \#(P_{un})
\label{EDUN}
\end{equation}
where $A_{un}$ denotes actual edges present between the union of neighbors of $w_i$ and $w_j$ and $P_{un}$ denotes the maximum possible between the union of neighbors.\\

The feature $SS$ captures mainly the degree of overlap of the neighborhoods of the word pairs, whereas $SPW$ and $SP$ indicate the distance between them in the DT network by considering and not considering the weight of the edges along the shortest path, respectively. The intuition behind taking these features is that if two words are distributionally very similar, there should be a short path between the two words via common neighbors. We observe in the DT network that, sometimes only the length of the shortest path is not enough to indicate the distributional similarity between two words; the average edge weight along the shortest path provides the hints of similarity between two words as well. This is the intuition behind proposing the measure $SPW$ along with $SP$.   
The last two proposed features, $ED_{in}$ and $ED_{un}$, capture the degree of closeness between the neighborhood of the word pair. 
Table \ref{samples} notes the values obtained for these network properties for sample pair of words for each relation type extracted from BLESS dataset. It is clearly seen that the $SP$, $ED{in}$ and $ED{un}$ values are higher for co-hyponymy pairs  compared to other relations and the other two features $SP$ and $SPW$ are comparatively lower, justifying the fact that co-hyponymy pairs are distributionally more similar and the words exist in close proximity in the DT network.

We now use these five features in different classifiers like SVM, Random Forest (as used in the baseline systems) to discriminate the co-hyponym word pairs from the word pairs having hypernymy or meronymy relation or any random pairs of word. 

\section{Experimental Results and Analysis}
As our main focus is classification of co-hyponymy relation, one of the key challenges has been to construct a dataset. Most of the gold standard datasets used for evaluation of the systems discriminating lexical relations, do not contain word pairs having co-hyponymy relation. We find two baseline systems~\cite{weeds2014learning,SANTUS16.455} where the authors use gold standard datasets which contain co-hyponymy pairs and they have done classification of co-hyponym pairs as well. 
We plan to evaluate our approach, by executing three experiments. In the first two experiments, we use the same experimental setup as well as the gold standard dataset of two baseline papers as used by the authors above. In the third experiment, we prepare our dataset from BLESS and do binary classification between co-hyponymy and other relations separately. 

\noindent{\bf Experiment 1:} In the first experiment, we directly use \boldmath$cohyponym_{BLESS}$, the gold standard dataset prepared by~\cite{weeds2014learning} from BLESS dataset~\cite{baroni2011we}. It contains 5,835 labelled pair of nouns, where for each BLESS concept, the co-hyponyms are considered as positive examples and the same total number of (and split evenly) hypernyms, meronyms and random words is taken as the negative examples. In addition to that, the order of 50\% of the pairs is reversed and duplicate pairs are disallowed. We use the same experimental setup of using SVM classifier with ten-fold cross validation as used by~\newcite{weeds2014learning} for this co-hyponymy classification task. ~\newcite{weeds2014learning} represent each word as positive
point wise mutual information (PPMI) based feature vector and then try to classify the relation between the given pair of words by feeding the word vectors to the classifier models using different vector operations. The details of the baselines as defined by ~\newcite{weeds2014learning} are presented in Table~\ref{Desc}.   
\begin{table}[!h]
\centering
\begin{tabular}{|>{\centering}p{1.4cm}|>{\centering}p{6cm}|}

      \hline
       \textbf{Baseline Model}&\textbf{Description}\tabularnewline \hline
       svmDIFF& A linear SVM trained on the vector difference\tabularnewline\hline  
	   svmMULT& A linear SVM trained on the pointwise product vector\tabularnewline \hline
svmADD&A linear SVM trained on the vector sum\tabularnewline \hline 
svmCAT&A linear SVM trained on the vector concatenation\tabularnewline  \hline
svmSING&A linear SVM trained on the vector of the second word in the given word pair\tabularnewline \hline
 knnDIFF& $k$ nearest neighbours (knn) trained on the vector difference \tabularnewline\hline
       cosineP& The relation between word pair holds if the cosine similarity of the word vectors is greater than some threshold $p$\tabularnewline\hline  
       linP&The relation between word pair holds if the lin similarity~\cite{lin1998automatic} of the word vectors is greater than some threshold $p$\tabularnewline\hline
       most freq & The most frequent label in the training data is assigned to every test point. \tabularnewline\hline

\end{tabular}

\caption{Descriptions of the baseline models as described in~\protect\cite{weeds2014learning}}
\label{Desc}
\end{table}

The performance  of our model along with these baselines is presented in Table~\ref{C_BLESS}. In the bottom part of Table~\ref{C_BLESS}, we present the result of our models where SVM classifier is used with each of the network features ($SS, SP, SPW, ED_{in}, ED_{un}$) separately. We try with using all five features together in a SVM classifier but it gives the same performance as using $SS$ only. We see that, instead of representing words as vectors and using several vector operations as features to SVM, simple network measures computed from Distributional Thesaurus Network lead to better or comparable performance. The network features are so strong that using any single feature, we achieve better performance compared to the supervised baselines (first 6 entries in Table~\ref{C_BLESS}) and the na\"{i}ve baseline of taking the most frequent label in the training data. On the other hand, we achieve comparable performance to the weakly supervised threshold based models (cosineP and linP) whereas for some features we beat those baselines gaining accuracy gain of 5\% with respect to the most competitive one.      

\begin{table}[!h]
\centering
\begin{tabular}{|>{\centering}p{2cm}|>{\centering}p{2cm}|>{\centering}p{2cm}|}

      \hline
       &\textbf{Model}&\textbf{Accuracy}\tabularnewline \hline
       \multirow{8}{*}{\textbf{Baselines}}&svmDIFF& 0.62\tabularnewline  
	   &svmMULT&0.39\tabularnewline 
       &svmADD&0.41\tabularnewline  
       &svmCAT&0.40\tabularnewline  
       &svmSING&0.40\tabularnewline 
       &knnDIFF&0.58\tabularnewline
       &cosineP&0.79\tabularnewline  
       &linP&0.78\tabularnewline
       &most freq & 0.61 \tabularnewline  
       \hline
	    \multirow{5}{*}{\textbf{Our models}}&svmSS&\textbf{0.84}\tabularnewline  
       &svmSP&0.83 \tabularnewline  
       &svmSPW&0.83 \tabularnewline  
	   &svm$ED_{in}$&0.78\tabularnewline  
	   &svm$ED_{un}$&0.76\tabularnewline \hline
\end{tabular}

\caption{Accuracy scores for $cohyponym_{BLESS}$ dataset of our model along with the models described in~\protect\cite{weeds2014learning}}
\label{C_BLESS}
\end{table}

\noindent{\bf Experiment 2:} In the second experiment, we use ROOT9 dataset, prepared by~\newcite{SANTUS16.455}. It contains 9600 labelled pairs randomly extracted from three datasets: EVALution~\cite{santus2015evalution}, Lenci/Benotto~\cite{benotto2015distributional} and BLESS~\cite{baroni2011we}. The dataset is evenly distributed among the three classes (hypernyms, co-hyponyms and random) and involves three types of parts of speech (noun, verb, adjective). The full dataset contains a total of 4,263 distinct terms  consisting of 2,380 nouns, 958 verbs and 972 adjectives. Here also, we use the same experimental setup of using Random Forest classifier with ten-fold cross validation as done by \citelanguageresource{santus2015evalution}. We have put all the five network measures as features to the classifier. We try with all the combinations of the five features and get the best performance when all of those features are used together. The performance  of our model along with the baselines are presented in Table~\ref{ROOT9}.
We see that in the binary classification task of Co-hyponym vs Random, we outperform all the state-of-the-art models in terms of F1 score whereas for Co-hyponym vs Hypernym classification task, our model beats the performance of most of the baseline models and produces comparable performance to the best models. 
Note that, using only five simple network measures as features we are able to get good performance, which leads to the fact that coming up with some  useful features intelligently can help in improving the performance of the otherwise difficult task of co-hyponymy detection. Investigating the DT network more deeply and coming up with some more sophisticated measures for co-hyponymy discrimination specially from hypernymy would definitely be the immediate future work.  

\begin{table}[!h]
\centering
\begin{tabular}{|>{\centering}p{3cm}|>{\centering}p{2cm}|>{\centering}p{2cm}|} 
\hline
      \textbf{Method}& \textbf{Co-Hyp vs Random}& \textbf{Co-Hyp vs Hyper} \tabularnewline \hline
       ROOT13& 97.4 & 94.3\tabularnewline \hline
	   ROOT9 & 97.8 & \textbf{95.7}\tabularnewline \hline
       -using SMO & 93.0 & 77.3\tabularnewline \hline
       -using Logistic & 95.3 & 78.7\tabularnewline \hline
       COSINE&79.4&69.8\tabularnewline \hline
       RANDOM13 & 51.4 & 50.1\tabularnewline \hline
        \hline
       Our Model& \textbf{99.0}& 87.0\tabularnewline \hline
\end{tabular}

\caption{Percentage F1 scores of our model along with the models described in~\protect\cite{SANTUS16.455} on a 10-fold cross validation for binary classification.  }
\label{ROOT9}
\vspace{-0.2cm}
\end{table}

\noindent{\bf Experiment 3:} The two experiments discussed so far show that using the proposed five network measures in classifiers gives better performance than the state-of-the-art models in the baseline datasets. Further, in order investigate the robustness of our approach, we create our own dataset extracted from BLESS~\cite{baroni2011we} for three binary classification tasks: Co-Hypo vs Hyper, Co-Hypo vs Mero, Co-Hypo Vs Random. For each of these tasks we have taken 1,000 randomly extracted pairs for positive instance (co-hyponymy pair) and 1,000 randomly extracted pairs for negative instance (hypernymy, meronymy and random pair, respectively). We have tried with both SVM and Random Forest classifiers with different combination of the proposed five features. Table~\ref{PE} presents the result of the best feature combination for both the classifiers for each of the binary classification task separately. We see that the performance of SVM classifier with only one feature structural similarity (SS) and Random Forest classifier with all the five features together provide good performance for all three binary classification tasks, consistent with the first two experiments. Note that, even though we get accuracy in the range of \textbf{0.86-0.97} while discriminating co-hyponym pairs from meronym or random pairs, we do not achieve highly accurate results when it comes to classification against hypernym pairs, indicating the fact that words having hypernymy relation and words having co-hyponymy relation may be having similar kind of neighborhood in the DT network, and further research is needed to discriminate between these using network measures only. 

\begin{table}[!h]
\centering
\begin{tabular}{|>{\centering}p{2cm}|>{\centering}p{2.5cm}|>{\centering}p{2.5cm}|}

      \hline
      \textbf{Classification} & \textbf{svmSS} & \textbf{random  forestALL}\tabularnewline
      \hline
       Co-Hyp vs Random & 0.96 & 0.97\tabularnewline \hline
	   Co-Hyp vs Mero & 0.86 & 0.89\tabularnewline  \hline
       Co-Hyp vs Hyper & 0.73 & 0.78\tabularnewline  \hline
\end{tabular}

\caption{Accuracy scores of on a 10-fold cross validation for binary classification using SVM and Random forest classifier.}
\label{PE}
\vspace{-0.5cm}
\end{table}

\section{Conclusion}
In this paper, we have proposed a supervised approach for discriminating co-hyponym pairs from hypernym, meronym and random pairs. We have introduced five symmetric complex network measures which can be used as features for the classifiers to detect co-hyponym pairs. By extensive experiments, we have shown that the proposed five features are strong enough to be fed into a classifier and beat the performance of most of the state-of-the-art models. Note that, applying distributional hypothesis to a corpus to build a Distributional Thesaurus (DT) network and computing small number of simple network measures is less computationally intensive compared to preparing vector representation of words. So in that sense this work contributes to an interesting finding that by applying complex network theory, we can devise an efficient supervised framework for co-hyponymy detection which performs better or at par in some cases, compared to the heavy-weight state-of-the-art models.     

The next immediate step is to use the proposed supervised features to guide in building unsupervised measures for co-hyponymy detection. In future, we plan to come up with some more sophisticated complex network measures like degree centrality, betweenness centrality etc. to be used for more accurate co-hyponymy detection. We also would like to investigate the possibilities of detecting hypernymy, meronymy relations with some asymmetric network measures. Finally, our broad objective is to build a general supervised and unsupervised framework based on complex network theory to detect different lexical relations from a given a corpus with high accuracy.


\begin{thebibliography}{}

\bibitem[\protect\citename{Antiqueira \bgroup et al.\egroup
  }2007]{antiqueira2007strong}
Antiqueira, L., Nunes, M. d. G.~V., Oliveira~Jr, O., and Costa, L. d.~F.
\newblock (2007).
\newblock Strong correlations between text quality and complex networks
  features.
\newblock {\em Physica A: Statistical Mechanics and its Applications},
  373:811--820.

\bibitem[\protect\citename{Baroni and Lenci}2011]{baroni2011we}
Baroni, M. and Lenci, A.
\newblock (2011).
\newblock How we blessed distributional semantic evaluation.
\newblock In {\em Proceedings of the GEMS 2011 Workshop on GEometrical Models
  of Natural Language Semantics}, pages 1--10. Association for Computational
  Linguistics.

\bibitem[\protect\citename{Benotto}2015]{benotto2015distributional}
Benotto, G.
\newblock (2015).
\newblock {\em Distributional Models for Semantic Relations: A Study on
  Hyponymy and Antonymy}.

\bibitem[\protect\citename{Berland and Charniak}1999]{berland1999finding}
Berland, M. and Charniak, E.
\newblock (1999).
\newblock Finding parts in very large corpora.
\newblock In {\em Proceedings of the 37th annual meeting of the Association for
  Computational Linguistics on Computational Linguistics}, pages 57--64.
  Association for Computational Linguistics.

\bibitem[\protect\citename{Cederberg and Widdows}2003]{cederberg2003using}
Cederberg, S. and Widdows, D.
\newblock (2003).
\newblock Using lsa and noun coordination information to improve the precision
  and recall of automatic hyponymy extraction.
\newblock In {\em Proceedings of the seventh conference on Natural language
  learning at HLT-NAACL 2003-Volume 4}, pages 111--118. Association for
  Computational Linguistics.

\bibitem[\protect\citename{Ferrer~i Cancho and Sol{\'e}}2001]{i2001small}
Ferrer~i Cancho, R. and Sol{\'e}, R.~V.
\newblock (2001).
\newblock The small world of human language.
\newblock {\em Proceedings of the Royal Society of London B: Biological
  Sciences}, 268(1482):2261--2265.

\bibitem[\protect\citename{Ferrer~i Cancho \bgroup et al.\egroup
  }2007]{ferrer2007spectral}
Ferrer~i Cancho, R., Capocci, A., and Caldarelli, G.
\newblock (2007).
\newblock Spectral methods cluster words of the same class in a syntactic
  dependency network.
\newblock {\em International Journal of Bifurcation and Chaos},
  17(07):2453--2463.

\bibitem[\protect\citename{Ferrer~i Cancho}2004]{i2004r}
Ferrer~i Cancho, R.
\newblock (2004).
\newblock R.; koehler, r.; sol{\'e}, rv patterns in syntactic dependency
  networks.
\newblock {\em Phys. Rev. E}, 69:32767.

\bibitem[\protect\citename{Firth}1957]{firth1957synopsis}
Firth, J.~R.
\newblock (1957).
\newblock A synopsis of linguistic theory, 1930-1955.
\newblock {\em Studies in linguistic analysis}.

\bibitem[\protect\citename{Fu \bgroup et al.\egroup }2014]{fu2014learning}
Fu, R., Guo, J., Qin, B., Che, W., Wang, H., and Liu, T.
\newblock (2014).
\newblock Learning semantic hierarchies via word embeddings.
\newblock In {\em ACL (1)}, pages 1199--1209.

\bibitem[\protect\citename{Geffet and Dagan}2005]{geffet2005distributional}
Geffet, M. and Dagan, I.
\newblock (2005).
\newblock The distributional inclusion hypotheses and lexical entailment.
\newblock In {\em Proceedings of the 43rd Annual Meeting on Association for
  Computational Linguistics}, pages 107--114. Association for Computational
  Linguistics.

\bibitem[\protect\citename{Girju \bgroup et al.\egroup
  }2006]{girju2006automatic}
Girju, R., Badulescu, A., and Moldovan, D.
\newblock (2006).
\newblock Automatic discovery of part-whole relations.
\newblock {\em Computational Linguistics}, 32(1):83--135.

\bibitem[\protect\citename{Goldberg and Orwant}2013]{goldberg2013dataset}
Goldberg, Y. and Orwant, J.
\newblock (2013).
\newblock A dataset of syntactic-ngrams over time from a very large corpus of
  english books.
\newblock In {\em Second Joint Conference on Lexical and Computational
  Semantics (* SEM)}, volume~1, pages 241--247.

\bibitem[\protect\citename{Hearst}1992]{hearst1992automatic}
Hearst, M.~A.
\newblock (1992).
\newblock Automatic acquisition of hyponyms from large text corpora.
\newblock In {\em Proceedings of the 14th conference on Computational
  linguistics-Volume 2}, pages 539--545. Association for Computational
  Linguistics.

\bibitem[\protect\citename{Kiela \bgroup et al.\egroup
  }2015]{kiela2015exploiting}
Kiela, D., Rimell, L., Vulic, I., and Clark, S.
\newblock (2015).
\newblock Exploiting image generality for lexical entailment detection.
\newblock In {\em Proceedings of the 53rd Annual Meeting of the Association for
  Computational Linguistics (ACL 2015)}, pages 119--124. ACL.

\bibitem[\protect\citename{Lin}1998]{lin1998automatic}
Lin, D.
\newblock (1998).
\newblock Automatic retrieval and clustering of similar words.
\newblock In {\em Proceedings of the 17th international conference on
  Computational linguistics-Volume 2}, pages 768--774. Association for
  Computational Linguistics.

\bibitem[\protect\citename{Miller}1995]{miller1995wordnet}
Miller, G.~A.
\newblock (1995).
\newblock Wordnet: a lexical database for english.
\newblock {\em Communications of the ACM}, 38(11):39--41.

\bibitem[\protect\citename{Morlane-Hond{\`e}re}2015]{morlane2015can}
Morlane-Hond{\`e}re, F.
\newblock (2015).
\newblock What can distributional semantic models tell us about part-of
  relations?
\newblock In {\em NetWordS}, pages 46--50.

\bibitem[\protect\citename{Nguyen \bgroup et al.\egroup
  }2017]{nguyen-EtAl:2017:EMNLP2017}
Nguyen, K.~A., K\"{o}per, M., Schulte~im Walde, S., and Vu, N.~T.
\newblock (2017).
\newblock Hierarchical embeddings for hypernymy detection and directionality.
\newblock In {\em Proceedings of the 2017 Conference on Empirical Methods in
  Natural Language Processing}, pages 233--243, Copenhagen, Denmark, September.
  Association for Computational Linguistics.

\bibitem[\protect\citename{Pantel and Pennacchiotti}2006]{pantel2006espresso}
Pantel, P. and Pennacchiotti, M.
\newblock (2006).
\newblock Espresso: Leveraging generic patterns for automatically harvesting
  semantic relations.
\newblock In {\em Proceedings of the 21st International Conference on
  Computational Linguistics and the 44th annual meeting of the Association for
  Computational Linguistics}, pages 113--120. Association for Computational
  Linguistics.

\bibitem[\protect\citename{Riedl and Biemann}2013]{riedl2013scaling}
Riedl, M. and Biemann, C.
\newblock (2013).
\newblock Scaling to large3 data: An efficient and effective method to compute
  distributional thesauri.
\newblock In {\em EMNLP}, pages 884--890.

\bibitem[\protect\citename{Roller and Erk}2016]{roller-erk:2016:EMNLP2016}
Roller, S. and Erk, K.
\newblock (2016).
\newblock Relations such as hypernymy: Identifying and exploiting hearst
  patterns in distributional vectors for lexical entailment.
\newblock In {\em Proceedings of the 2016 Conference on Empirical Methods in
  Natural Language Processing}, pages 2163--2172, Austin, Texas, November.
  Association for Computational Linguistics.

\bibitem[\protect\citename{Roller \bgroup et al.\egroup
  }2014]{roller2014inclusive}
Roller, S., Erk, K., and Boleda, G.
\newblock (2014).
\newblock Inclusive yet selective: Supervised distributional hypernymy
  detection.
\newblock In {\em COLING}, pages 1025--1036.

\bibitem[\protect\citename{Santus \bgroup et al.\egroup
  }2014]{santus2014chasing}
Santus, E., Lenci, A., Lu, Q., and Im~Walde, S.~S.
\newblock (2014).
\newblock Chasing hypernyms in vector spaces with entropy.
\newblock In {\em EACL}, pages 38--42.

\bibitem[\protect\citename{Santus \bgroup et al.\egroup
  }2015]{santus2015evalution}
Santus, E., Yung, F., Lenci, A., and Huang, C.-R.
\newblock (2015).
\newblock Evalution 1.0: an evolving semantic dataset for training and
  evaluation of distributional semantic models.
\newblock In {\em Proceedings of the 4th Workshop on Linked Data in Linguistics
  (LDL-2015)}, pages 64--69.

\bibitem[\protect\citename{Santus \bgroup et al.\egroup }2016]{SANTUS16.455}
Santus, E., Lenci, A., Chiu, T.-S., Lu, Q., and Huang, C.-R.
\newblock (2016).
\newblock Nine features in a random forest to learn taxonomical semantic
  relations.
\newblock In {\em Proceedings of the Tenth International Conference on Language
  Resources and Evaluation (LREC 2016)}, Paris, France, may. European Language
  Resources Association (ELRA).

\bibitem[\protect\citename{Shwartz \bgroup et al.\egroup
  }2017]{shwartz-santus-schlechtweg:2017:EACLlong}
Shwartz, V., Santus, E., and Schlechtweg, D.
\newblock (2017).
\newblock Hypernyms under siege: Linguistically-motivated artillery for
  hypernymy detection.
\newblock In {\em Proceedings of the 15th Conference of the European Chapter of
  the Association for Computational Linguistics: Volume 1, Long Papers}, pages
  65--75, Valencia, Spain, April. Association for Computational Linguistics.

\bibitem[\protect\citename{Weeds \bgroup et al.\egroup
  }2014]{weeds2014learning}
Weeds, J., Clarke, D., Reffin, J., Weir, D., and Keller, B.
\newblock (2014).
\newblock Learning to distinguish hypernyms and co-hyponyms.
\newblock In {\em Proceedings of COLING 2014, the 25th International Conference
  on Computational Linguistics: Technical Papers}, pages 2249--2259. Dublin
  City University and Association for Computational Linguistics.

\bibitem[\protect\citename{Yamada \bgroup et al.\egroup
  }2009]{yamada2009hypernym}
Yamada, I., Torisawa, K., Kazama, J., Kuroda, K., Murata, M., De~Saeger, S.,
  Bond, F., and Sumida, A.
\newblock (2009).
\newblock Hypernym discovery based on distributional similarity and
  hierarchical structures.
\newblock In {\em Proceedings of the 2009 Conference on Empirical Methods in
  Natural Language Processing: Volume 2-Volume 2}, pages 929--937. Association
  for Computational Linguistics.

\bibitem[\protect\citename{Yu \bgroup et al.\egroup }2015]{yu2015learning}
Yu, Z., Wang, H., Lin, X., and Wang, M.
\newblock (2015).
\newblock Learning term embeddings for hypernymy identification.
\newblock In {\em IJCAI}, pages 1390--1397.

\end{thebibliography}
\end{document}